\documentclass[letterpaper, 10 pt, conference]{ieeeconf}  

\IEEEoverridecommandlockouts                              

\usepackage{amsmath} 
\usepackage{amssymb}  
\usepackage{amsfonts}
\usepackage{dsfont}
\usepackage{graphicx}
\usepackage{algorithm}
\usepackage[noend]{algorithmic}
\usepackage{psfrag,graphicx,epsfig}
\usepackage{epstopdf}
\usepackage{xspace}
\usepackage{subfig}
\usepackage{float}
\usepackage{placeins}
\usepackage{multirow}
\usepackage{pgf,tikz}
\usepackage{nowidow}
\usepackage{lineno}
\usepackage{xcolor}
\newcommand{\fig}[1]{Fig.~\ref{#1}}
\newcommand{\tab}[1]{Table~\ref{#1}}
\newcommand{\eq}[1]{(\ref{#1})}

\DeclareMathOperator*{\argmin}{arg\,min}

\usepackage{siunitx}
\usepackage{color}
\usepackage{flushend}
\usepackage{lineno}
\usepackage{mathtools}
\allowdisplaybreaks

\title{\LARGE \bf
Simultaneous Contact-Rich Grasping and Locomotion via Distributed Optimization Enabling Free-Climbing for Multi-Limbed Robots}

\author{Yuki Shirai$^{\dagger}$, Xuan Lin$^{\dagger}$, Alexander Schperberg$^{\dagger}$, Yusuke Tanaka$^{\dagger}$, \\ Hayato Kato$^{\dagger}$,  Varit Vichathorn$^{\dagger}$, and Dennis Hong$^{\dagger}$%
\thanks{$^{\dagger}$ All authors are with the Department of Mechanical and Aerospace Engineering, University of California, Los Angeles, CA, USA 90095 {\tt\small \{yukishirai4869, maynight, aschperberg28, yusuketanaka, hayatokato, omezzvct, dennishong\}@g.ucla.edu}}}%

\begin{document}
\bstctlcite{IEEEexample:BSTcontrol}

\maketitle


\begin{abstract}
While motion planning of locomotion for legged robots has shown great success, motion planning for legged robots with dexterous multi-finger grasping  is not mature yet. 
We present an efficient motion planning framework for simultaneously solving locomotion (e.g., centroidal dynamics), grasping (e.g., patch contact), and contact (e.g., gait) problems.
To accelerate the planning process, we propose distributed optimization frameworks based on Alternating Direction Methods of Multipliers (ADMM) to 
solve the original large-scale Mixed-Integer NonLinear Programming (MINLP).
The resulting frameworks use Mixed-Integer Quadratic Programming (MIQP) to solve contact and NonLinear Programming (NLP) to solve nonlinear dynamics, which are more computationally tractable and less sensitive to parameters.
%
%
Also, 
we explicitly enforce patch contact constraints from limit surfaces with micro-spine grippers. We demonstrate our proposed framework in the hardware experiments, showing that the multi-limbed robot is able to realize various motions including free-climbing at a slope angle $\mathbf{\ang{45}}$ with a much shorter planning time.
\end{abstract}

\section{Introduction}\label{sec:introduction}



While legged robots have shown remarkable success in locomotion tasks such as running, 
legged robots with  dexterous manipulation skills, defined as limbed robots in this paper, are relatively unexplored. There are a number of promising applications for limbed robots such as manipulating balls \cite{Circus}, 
sitting \cite{doi:10.1080/01691864.2012.686345}, bobbin rolling \cite{9359455}, pushing heavy objects \cite{8957267}, 
stair-climbing \cite{9478184}, and free-climbing  \cite{7989643}.
In this paper, we focus on free-climbing tasks of limbed robots. Free-climbing capabilities would be useful for planetary exploration, inspection, and so on. 
These tasks cannot be done by traditional legged robots by dismissing these problems as locomotion tasks. 
All of those previous works consider coupling effects between body stability of legged robots and frictional interaction of manipulators to some extent.
To implement those capabilities in a real limbed robot, a variety of physical constraints need to be considered.
%
Thus, motion planning plays a key role to generate physically feasible trajectories of limbed robots for those non-trivial tasks.

However, motion planning of limbed robots for such tasks is challenging. First, motion planning can be quite complicated because planners need to solve trajectories of legged robots,  manipulators, or grippers together, which leads to NLP if motion planning is formulated as an optimization problem. Also, it is difficult to identify contact sequences prior to motion planning if the task is complicated such as free-climbing. Thus, it is required to consider locomotion, manipulation, and contacts together for generating trajectories, which results in computationally intractable MINLP.  

Another problem arises when limbed robots interact with environments using patch contacts. With patch contacts, limbed robots can effectively increase friction forces and use friction torques generated on the patch, which is useful for manipulation tasks and even free-climbing tasks. However, multi-finger patch contacts are not discussed yet in previous motion planning works for limbed robots. 

\begin{figure}
    \centering
    \includegraphics[width=0.48\textwidth]{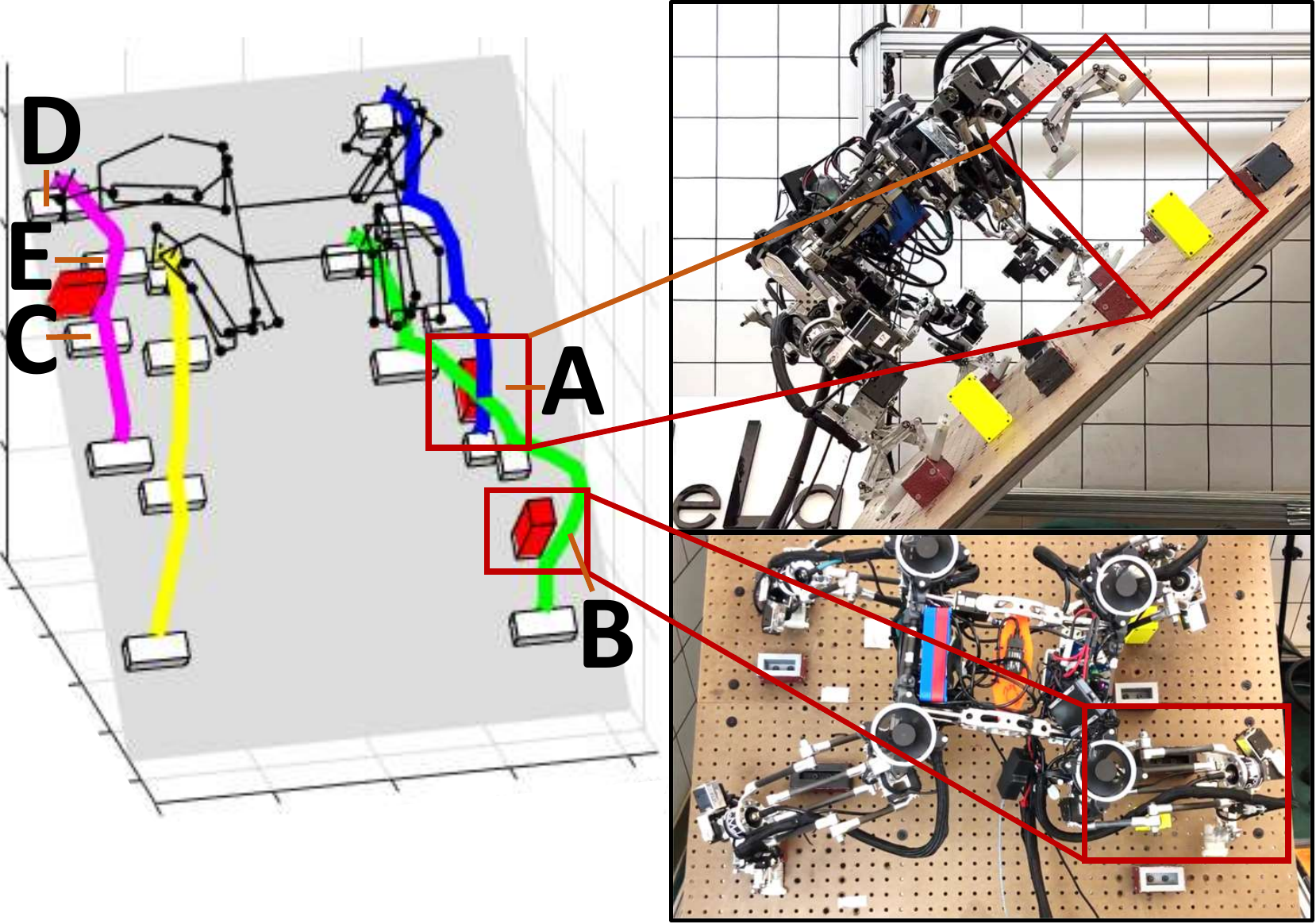} %
    \caption{We consider motion planning of multi-limbed robots for free-climbing. Our proposed framework efficiently generates trajectories for multi-limbed robots equipped with multi-finger grippers while considering locomotion, grasping, and contacts. The left figure shows trajectories of one of the fingers of each gripper while the robot avoids obstacles. The trajectories around A-E are discussed in Sec~\ref{result_tra_to}. The right figure shows that our real four-limbed robot executes our planned trajectory.}
    \label{fig:collision}
\end{figure}


In this paper, we propose a motion planning algorithm that efficiently solves locomotion, grasping, and contact dynamics together. We show that
the resulting framework is computationally more tractable and less sensitive to parameters. 
Also, we explicitly discuss the patch contact constraints with micro-spine grippers, which enables the algorithm to  realize dexterous multi-finger tasks. 
Our proposed motion planning is validated on a 9.6 kg four-limbed robot with spine grippers for  free-climbing. 
To the best of our knowledge, it is one of the first works that demonstrate dexterous multi-finger grasping enabling free-climbing on a real multi-limbed robot.

This paper presents the following contributions:
\begin{enumerate}
\item We present an optimization-based motion planning framework that simultaneously solves constraints from locomotion, grasping and contact dynamics. 
\item We accelerate the entire optimization process by formulating the problem as a distributed optimization.
\item We explicitly formulate patch contact constraints for micro-spine grippers.
\item We validate our framework in hardware experiments. 
\end{enumerate}

\section{Related Work}\label{sec:related_work}

Motion planning based on Contact-Implicit Trajectory Optimization (CITO) has been studied in locomotion and manipulation literature \cite{doi:10.1177/0278364913506757, yuki2021pivot, 8740889, 8141917}. Some of those works use complementarity constraints and other works use integer constraints to model contact. 
For both approaches, however, the computational complexity increases and it can be challenging to find feasible solutions as the number of discrete modes increases.
Thus, 
many works solve the approximated MINLP. 
In \cite{8283570, 9166536}, the authors use NLP with phase-based formulations. 
One drawback is that the order of phases cannot be changed, which is undesirable for motion planning for limbed robots since the number of phases limbed robot planners consider is large and it can lead to infeasible solutions.
The authors in \cite{8957267, 9113247} use continuous formulation in NLP to represent discrete terrain. This work considers fully discrete environments for free-climbing tasks so we cannot use these techniques. 
In \cite{9478184, nguyen2021contact}, the authors decouple the MINLP problem as \textit{sequential} sub-problems (i.e., hierarchical planning).
%
However, such a formulation cannot guarantee that the entire planning process is feasible since it does not in general consider all coupling constraints among sub-problems. 



Our work is inspired by distributed optimization such as ADMM \cite{MAL-016}, which has gathered attention for large-scale optimization problems \cite{8793878, 9147887, aydinoglu2021realtime, 9620665}. In \cite{8793878, 9147887}, the authors introduce an ADMM-based framework to reason centroidal and whole-body dynamics. This work does not consider whole-body dynamics but considers contact dynamics from grippers and discrete constraints. ADMM is also employed in Model Predictive Control (MPC)  for linear complementarity problem \cite{aydinoglu2021realtime}. The work in \cite{9620665} proposed an ADMM-based framework for CITO. We instead consider nonlinear centroidal dynamics and propose a specific splitting scheme for motion planning of limbed robots.
\section{Patch Contact Model with Micro-Spines}\label{sec:patch}


In this section, we discuss the patch contact model used in our planner. 
We extend the previous works \cite{doi:10.1177/027836499601500603, wang2017design, 8416785} and explicitly incorporate the limit surface in our proposed  planner. As there is a normal force acting on the patch, the limit surface is composed of the gripper force failure model
and the friction failure model. The total available reaction wrench $\mathbf{w} = [f^x, f^y, f^z, \tau^x,  \tau^y, \tau^z]$ lives in a Minkowski sum of those two models described by:
%
\begin{equation}
    \mathbf{w} \in \mathcal{W}, \ 
    \mathcal{W} = \{\mathbf{w_\text{fr}} + \mathbf{w_\text{sp}}| \mathbf{w_\text{fr}} \in \mathcal{W}_\text{fr}, \mathbf{w_\text{sp}} \in \mathcal{W}_\text{sp} \}
    \label{minkosum}
\end{equation}
where we define $z$-axis as the direction of normal forces and $x$- and $y$-axis consist $xy$ plane where shear forces exist (e.g., see $\Sigma_{c = 3}$ in \fig{fig:opt_figure}).
$\mathbf{{w}_\text{fr}} = [f_{\text{fr}}^x,  f_{\text{fr}}^y, f_{\text{fr}}^z,  \tau_{\text{fr}}^x,  \tau_{\text{fr}}^y, \tau_{\text{fr}}^z]$ is the friction wrench and 
$\mathbf{{w}_\text{sp}} = [f_{\text{sp}}^x,  f_{\text{sp}}^y, f_{\text{sp}}^z,  \tau_{\text{sp}}^x,  \tau_{\text{sp}}^y, \tau_{\text{sp}}^z]$ is the wrench that can be supported by the micro-spines with the zero normal force.
$\mathcal{W}_\text{fr}, \mathcal{W}_\text{sp}$ represent a frictional limit surface and a limit surface from micro-spines, respectively. 


\subsection{Frictional Limit Surface}\label{section_ls}
Previous literature \cite{doi:10.1177/027836499601500603}  models the friction wrench failure model as a simple 4D ellipsoid on $[f_{\text{fr}}^x, f_{\text{fr}}^y, f_{\text{fr}}^z,  \tau_{\text{fr}}^z]$ as follows:
\begin{equation}
    \begin{split}
    \mathcal{W}_\text{fr} = \{\mathbf{ w_\text{fr}} \in \mathbb{R}^{6} | \frac{(f_{\text{fr}}^x)^{2} + (f_{\text{fr}}^y)^{2}}{(\mu  f_{\text{fr}}^z)^{2}} + \frac{ (\tau_{\text{fr}}^z)^2}{(k \mu f_{\text{fr}}^z)^{2}} \leq 1, \\ 0 \leq f_\text{fr}^z \leq f_\text{max}, \tau_{\text{fr}}^x=\tau_{\text{fr}}^y=0\}
    \label{limit_surface_full}
\end{split}
\end{equation}
such that $\mu$ is the coefficient of friction, $k$ is an integration constant, $f_\text{max}$ is the upper bound of $f_\text{fr}^z$. This work assumes that fingers make circular patch contact under uniform pressure distribution and thus we use $k=0.67r_p$ where $r_p$ is the radius of contact \cite{doi:10.1177/027836499601500603}.

\subsection{Limit Surface of Micro-Spines}\label{section_microspine}
The authors in \cite{wang2017design} 
 construct the limit surface for spine grippers of any contact angles with the constant $\mu$.
 If the contact angle $\phi$ in \fig{fig:opt_figure} begins to vary, $\mu$  may become a function of $\phi$ and requires an independent model. In practice, building such a model requires a large amount of data.

This work simplifies the model by assuming that the patch is always perpendicular to the surface during contact (i.e., $\phi = \frac{\pi}{2}$).
Therefore, $\mu$ is  constant. We also make the assumption that the moment $\tau_{\text{fr}}^x, \tau_{\text{fr}}^y$  are negligible since the size of the patch is relatively small. However, the moment $\tau_{\text{fr}}^z$  cannot be neglected according to our test data. Thus, we impose a following 3D limit surface on $[f_{\text{sp}}^x, f_{\text{sp}}^y,  \tau_{\text{sp}}^z]$:
\begin{equation}
    \begin{split}
    \mathcal{W}_\text{sp} = \{\mathbf{ w_\text{sp}} \in \mathbb{R}^{6} | -f_{\text{max}}^i\leq f_{\text{sp}}^i \leq  f_{\text{max}}^i, i  = \{x, y\},\\ f_{\text{sp}}^z = 0, 
    \tau_{\text{sp}}^x=\tau_{\text{sp}}^y=0, -\tau_{\text{max}}^z\leq \tau_{\text{sp}}^z \leq  \tau_{\text{max}}^z\}
    \label{limit_surface_spine}
\end{split}
\end{equation}
where $f^i_\text{max}, \tau^i_\text{max}$ represent the upper bound of each wrench.  
 
\subsection{Limit Surface for Two-Finger Micro-Spine Grippers}
For the two-finger gripper used by our robot, each finger is equipped with a micro-spine patch with the total available  variables $\mathbf{w}_1 = [f_{\text{1}}^x, f_{\text{1}}^y, f_{\text{1}}^z,  \tau_{\text{1}}^z] \in \mathcal{W}$ and $\mathbf{w}_2 = [f_{\text{2}}^x, f_{\text{2}}^y, f_{\text{2}}^z,  \tau_{\text{2}}^z] \in \mathcal{W}$, where $\mathbf{w}_1$ is for finger 1 and $\mathbf{w}_2$ is for finger 2. Since the rotational motion along $z$-axis for finger 1 and finger 2 are same and the linear motion along $x$- and $y$-axis are same (see Sec~\ref{assumption}), we assume that the loading shear force and moment between two contact patches are identical:
\begin{equation}
    f_{\text{1}}^i = f_{\text{2}}^i, i  = \{x, y\},
    \tau_{\text{1}}^z =\tau_{\text{2}}^z
\end{equation}
This can be justified as there cannot be an additional twisting moment along $z$-axis from the object being grasped. 
\section{Simultaneous Contact-Rich Grasping and Locomotion Trajectory Optimization}\label{sec:robust_to}

In this section, we present our proposed optimization formulation which simultaneously solves grasping and locomotion while considering discrete dynamics such as gait sequence. Then, we derive the computationally tractable formulation of our proposed formulation based on ADMM.

\begin{table}[]
    \centering
        \caption{Notation of variables. C or B indicates the variable is continuous or binary variables, respectively. In $\Sigma$ column, we indicate the frame of variables. Subscript $t$ indicates time-step.}
\begin{tabular}{|c|c|c|c|c|}
\hline Name & Description & Size  & C/B & $\Sigma$ \\
\hline $\mathbf{r}_t$ & body position & $\mathbb{R}^3$ & C & $W$ \\
 $\boldsymbol{\theta}_t$ & body orientation & $\mathbb{R}^3$ & C & $W$ \\
 $\mathbf{p}^i_t$ & $i$-th finger position & $\mathbb{R}^3$ & C & $W$ \\
 $\mathbf{q}^i_t$ & $i$-th finger orientation & $\mathbb{R}^3$ & C & $W$ \\
 $\mathbf{d}^i_t$ & $l$-th gripper distance between fingers & $\mathbb{R}^3$ & C & $W$ \\
 $\boldsymbol{\lambda}^i_t$ & $i$-th finger reaction force & $\mathbb{R}^3$ & C & $W$ \\
 $\boldsymbol{\tau}^i_t$ & $i$-th finger reaction moment & $\mathbb{R}^3$ & C & $W$ \\
 $\mathbf{f}^{i, c}_t$ & $i$-th finger local force at $\mathcal{C}_c$ & $\mathbb{R}^3$ & C & $c$ \\
 $\mathbf{m}^{i, c}_t$ & $i$-th finger local moment at $\mathcal{C}_c$ & $\mathbb{R}^3$ & C & $c$ \\
\hline $\alpha^{i, c}_t$ & $i$-th finger contact at $\mathcal{C}_c$ & $\mathbb{Z}^1$ & B &  \\
 $\beta^{i, v, h}_t$ & $i$-th finger collision to $h$-th face of $\mathcal{V}_v$ & $\mathbb{Z}^1$ & B &  \\
  $\gamma^{i, c}_t$ & direction of $z$-element of  $m^{i, c}_t$ & $\mathbb{Z}^1$ & B &  \\ \hline
\end{tabular}
    \label{tab:my_label}
\end{table}

\subsection{Preliminary}\label{assumption}
\begin{figure}
    \centering
    \includegraphics[width=0.27\textwidth]{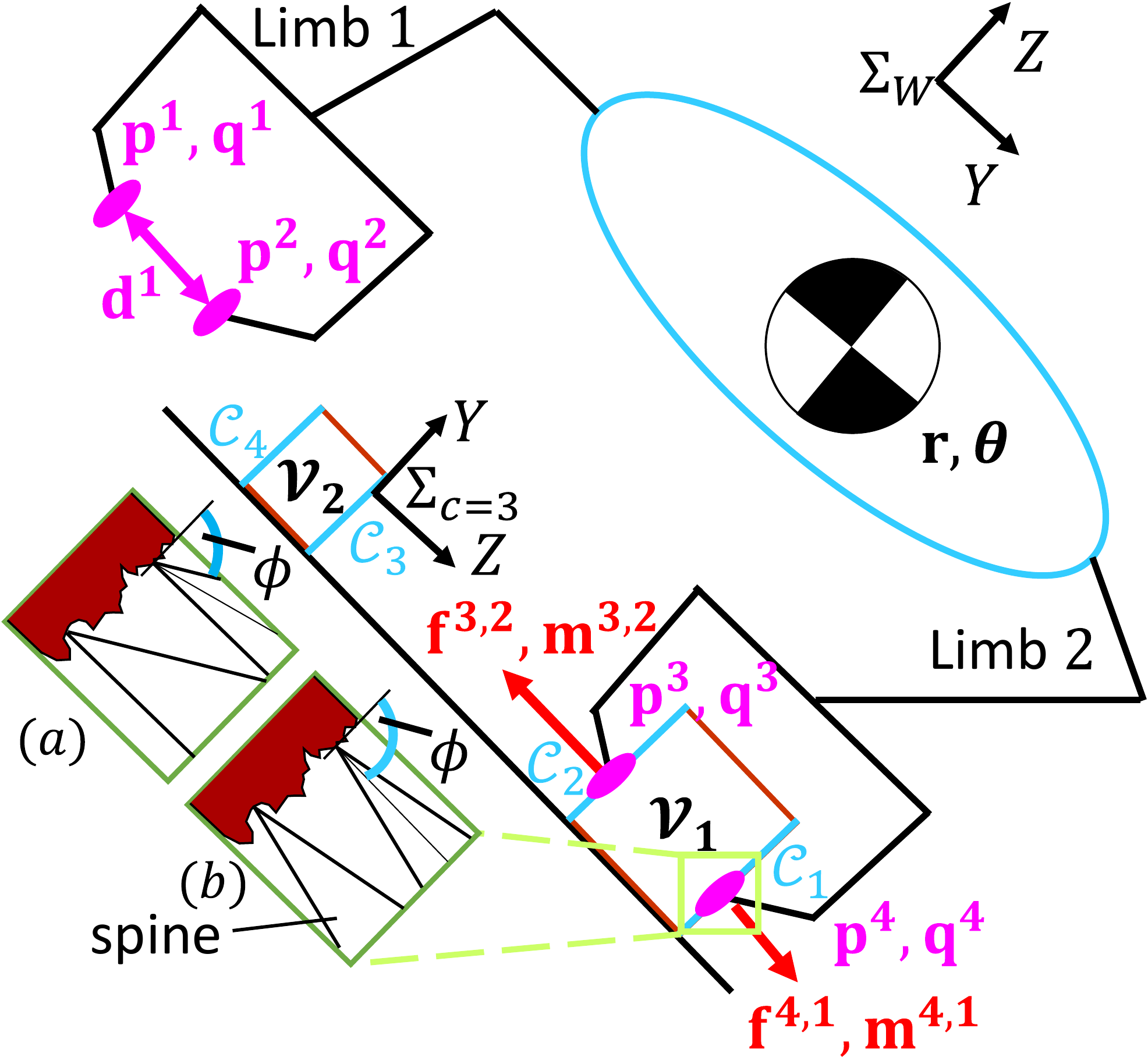} %
    \caption{Mathematical model of a multi-limbed robot. In this example, $n_f = 4, n_l = 2, C = 4, V = 2$. We also visualize two examples of frames $\Sigma_W, \Sigma_{c=3}$ where $z$-axis of $\Sigma_W$ is perpendicular to the ground and $z$-axis of the local frame $\Sigma_{c=3}$ is perpendicular to the face of the climbing hold and along this axis we have non-zero $z$ element of $m$. The spine makes contact with the contact angle (a): $\phi = \frac{\pi}{3}$ and (b): $\phi = \frac{\pi}{2}$.}
    \label{fig:opt_figure}
\end{figure}
Here, we show our assumptions in our planner:
\begin{enumerate}
\item Each finger makes patch contact and follows two different frictional models discussed in Sec~\ref{sec:patch}.
\item The environment consists of rigid static climbing holds whose geometry is modeled as cuboids.
\item Paired fingers align along the normal direction of fingertips. 
Paired fingers are in parallel and  rotate only along  $z$-axis in the world frame $\Sigma_W$.
%
\end{enumerate}
We define the variables in \tab{tab:my_label} and \fig{fig:opt_figure}. 
We also denote constants as follows.
$N, n_f, n_l, C$ or $V$ represent the time horizon, the total number of fingers, the total number of limbs, the total number of graspable regions, or the total number of obstacles, respectively. 
We denote $\mathcal{C}_c$ as the $c$-th graspable region, associated with a local frame $\Sigma_{c}$. 
Each obstacle $\mathcal{V}_v$ has $n_v$ faces associated with a local face frame $\Sigma_{v_h}, h = 1, \ldots, n_v$.
For any arbitrary vector $\mathbf{a}$, the notation $\|\mathbf{a}\|^2_{A}$ means a quadratic term with a positive-semi-definite matrix $A$. We define the coordinate transformation from frame $\Sigma_A$ to $\Sigma_B$ as $^A_BT$. We denote $X \Longrightarrow Y$ as a conditional constraint and implement it using a big-M formulation. 

\subsection{Optimal Control Problem for Grasping and Locomotion}
We propose the optimal control problem in \eq{equation_control}.
Our planner finds the optimal trajectory of body pose, limb poses, and wrenches, subject to limb and gripper kinematics,  centroidal dynamics, bound of variables, gait, faces to grasp, collision-avoidance, and our proposed limit surface constraints. 
%
\begin{equation}
\textbf{P1: }\min_{\mathbf{x, u, y, z}} \sum_{t=0}^{N-1} J(\mathbf{x}_t, \mathbf{u}_t, \mathbf{y}_t, \mathbf{z}_t) 
\label{equation_control}
\end{equation}
subject to:
\begin{enumerate}
\item For time-step $t=0, \ldots, N-1$
\begin{itemize}
    \item Bounds of decision variables \eq{bounds_eq}.
    \item Centroidal dynamics \eq{Centroidal_eq}.
    \item For fingers $i=1, \ldots, n_f$:
    \begin{itemize}
        \item Kinematics \eq{kinematics_eq}.
        \item For graspable regions $c=1, \ldots, C$:
        \begin{itemize}
            \item Contact constraints \eq{contact_eq}.
            \item Wrench transformation \eq{trans}.
        \end{itemize}
        \item For obstacles $v=1, \ldots, V$:
        \begin{itemize}
            \item Collision-avoidance \eq{collision_eq}.
        \end{itemize}
    \end{itemize}
\end{itemize}
    \item Terminal state constraints.
\end{enumerate}
We define $\mathbf{x}_t = [\mathbf{r}_t^\top, \boldsymbol{\theta}_t^\top, \dot{\mathbf{r}}_t^\top, \dot{\boldsymbol{\theta}}_t^\top, \mathbf{p}_t^{i\top}, \mathbf{q}_t^{i\top}, \mathbf{d}_t^{l\top}, \forall  i, l]^\top$, $\mathbf{u}_t = [\boldsymbol{\lambda}_t^{i\top}, \boldsymbol{\tau}_t^{i\top}, \forall i]^\top$, $\mathbf{y}_t = [\mathbf{f}_t^{i,c\top}, \mathbf{m}_t^{i,c\top}, \forall i, c]^\top$, and $\mathbf{z}_t = [\alpha_t^{i, c\top}, \beta_t^{i, v, h\top}, \gamma_t^{i, c\top}, \forall i, c, v, h]^\top$, where $i = 1, \ldots, n_f$, $l = 1, \ldots, n_l$, $c = 1, \ldots, C$, $v=1, \ldots, V$, $h=1, \ldots, n_v$. 


\subsubsection{Cost Function and Bounds}
Our cost function is:
\begin{equation}
\begin{aligned}
    J =  \|\mathbf{x}_t - \mathbf{x}_g\|^2_{Q} +\|\mathbf{u}_t\|^2_{R} + \boldsymbol{\zeta}^\top \mathbf{x}_t +\\ \sum_{l=1}^{n_l}\boldsymbol{\xi}_l^\top(\mathbf{p}_t^{2l}-\mathbf{p}_t^{2l-1}) +\|\mathbf{z}_{t+1} - \mathbf{z}_t\|^2_{S}
\end{aligned}
    \label{cost_actual}
\end{equation}
 The first term is the cost between the current state and the terminal state $\mathbf{x}_g$. The second term is the control effort cost. We aim to lift each limb as high as possible  since potential hazards (e.g., obstacles) can exist near terrain. However, this capability has not been realized well. In \cite{doi:10.1177/0278364913506757, 8740889}, the generated swing height is almost zero unless the authors give the reference trajectory.
 Hence,   by assigning a negative  value for elements of $\boldsymbol{\zeta} \in \mathbb{R}^{n_x}$ associated with limb heights in $\mathbf{x}_t$, our planner can swing limbs with reasonable heights.
 
 We observe that the distance between the surface of the climbing hold and each finger when the robot release the fingers needs to be long enough. Otherwise, due to an imperfect position controller, the finger can stick to the climbing hold, resulting in the failure of releasing fingers.
Hence, we maximize the distance between paired fingers with a negative  value for elements of each  $\boldsymbol{\xi}_l \in \mathbb{R}^3$,  
which indirectly increases the distance between the graspable region and the finger once the fingers release. 

We  observe that CITO randomly switches the discrete modes (e.g., contact on-off), which could lead to instability. 
 By assigning a quadratic term for $\mathbf{z}_t, \mathbf{z}_{t+1}$ associated with the mode we do not want to switch frequently, 
the fifth term in  \eq{cost_actual} prevents  mode changes between $t$ and $t+1$.
 
We bound the range of desicion varibles as follows:
\begin{equation}
    \mathbf{x}_t \in \mathcal{X}, \mathbf{u}_t \in \mathcal{U}, \mathbf{y}_t \in \mathcal{Y}, \mathbf{z}_t \in \mathcal{Z}
    \label{bounds_eq}
\end{equation}
where $\mathcal{X} \subseteq \mathbb{R}^{n_{x}}$, $\mathcal{U} \subseteq \mathbb{R}^{n_{u}}$, and $\mathcal{Y} \subseteq \mathbb{R}^{n_{y}}$ are convex polytopes consisting of a finite number of linear inequality constraints.  $\mathcal{Z} \subseteq \{0, 1\}^{n_{z}}$ shows range of binary variables. 

\subsubsection{Centroidal Dynamics}
The dynamics is given by:
\begin{subequations}
\begin{flalign}
    M \Ddot{\mathbf{r}}_t = \sum_{i=1}^{n_f} \boldsymbol{\lambda}_t^i + M\mathbf{g}\label{force_dynamics}\\
    I \dot{\boldsymbol{\omega}}_t + \boldsymbol{\omega}_t \times I\boldsymbol{\omega}_t = \sum_{i=1}^{n_f} \left(\mathbf{r}_t -\mathbf{p}_t^i\right) \times \boldsymbol{\lambda}_t^i + \boldsymbol{\tau}_t^i\label{moment_dynamics}
\end{flalign}
\label{Centroidal_eq}
\end{subequations}
where $M, I$ represent the mass and inertia of the robot. $\mathbf{g} \in \mathbb{R}^{3}$ is the gravity acceleration, and $\boldsymbol{\omega}_t$ is the angular velocity from $\boldsymbol{\theta}_t$ \cite{8283570}. This work explicitly considers $\boldsymbol{\tau}_t^i$ to capture the effect of patch contacts. For implementation, we use the explicit-Euler method with time interval $dt$.

\subsubsection{Kinematics}
Our kinematics constraints are as follows:
\begin{subequations}
\begin{flalign}
|R(\boldsymbol{\theta}_t)(\mathbf{p}_t^i-\mathbf{r}_t) - \mathbf{a}^i| \leq \mathbf{b}^i, 
|\mathbf{q}_t^i-\mathbf{c}^i-\boldsymbol{\theta}_t| \leq \mathbf{d}^i \label{kin_rot}\\
\mathbf{p}_t^{2l} = \mathbf{d}_t^l + \mathbf{p}_t^{2l-1}, \mathbf{q}_t^{2l} = \mathbf{q}_t^{2l-1} \label{kin_fingers}
\end{flalign}
\label{kinematics_eq}
\end{subequations}
$R(\boldsymbol{\theta}_t)$ is the rotation matrix from $\Sigma_W$ to $\Sigma_B$ where $\Sigma_B$ is the body frame. $\mathbf{a}^i, \mathbf{b}^i$ are the nominal position and acceptable range from the nominal position of $i$-th finger in $\Sigma_B$. 
$\mathbf{c}^i, \mathbf{d}^i$  represent the nominal orientation and acceptable range from the nominal orientation of $i$-th finger. 
In \eq{kin_fingers}, one of the paired finger positions is determined by another finger position and $\mathbf{d}_t^l$. Since fingers on the same gripper are parallel, we set the orientation of those paired fingers as same. 
Later, we use \eq{kinematics_eq} in  MIQP  and thus  conservatively approximate \eq{kinematics_eq} by linearizing $R(\boldsymbol{\theta}_t)$ at a certain angle.
 
\subsubsection{Contact Constraints}
Contact dynamics is inherently discrete phenomenon and thus it can be given by:
\begin{subequations}
\begin{flalign}
\alpha_t^{i, c} = 1 &\Longrightarrow
\left \{
\begin{aligned}
\mathbf{f}_t^{i, c}, \mathbf{m}_t^{i, c} \in \mathcal{W}\left(\mu_c, k_c\right),\\ \dot{\mathbf{p}}_t^i, \dot{\mathbf{q}}_t^i  = \mathbf{0}, 
{^{c}_WT} \left(\mathbf{p}_t^i, \mathbf{q}_t^i\right) \in \mathcal{C}_c
\end{aligned}
\right\}\label{contactA}
\\
\alpha_t^{i, c} = 0 &\Longrightarrow \mathbf{f}_t^{i, c}, \mathbf{m}_t^{i, c} = \mathbf{0}\label{contactB}\\
\sum_{c=1}^{C} \alpha_t^{i, c} &\leq 1 \label{sum_contact}
\end{flalign}
\label{contact_eq}
\end{subequations}
$\mu_c, k_c$ are parameters from \eq{limit_surface_full} defined in $\mathcal{C}_c$.
The constraints in \eq{contactA} mean that if the finger makes contact on $\mathcal{C}_c$, the local wrench $\mathbf{f}_t^{i, c}, \mathbf{m}_t^{i, c}$ needs to follow the patch constraints in \eq{minkosum} and the finger does not move. 
${^{c}_WT} \left(\mathbf{p}_t^i, \mathbf{q}_t^i\right)$ represents the $\mathbf{p}_t^i, \mathbf{q}_t^i$ in $\Sigma_{c}$.
If the finger is in the air, \eq{contactB} means that the local wrench is zero. Because the finger can only make contact on one of the graspable regions, \eq{sum_contact} does not allow the finger to make more than one contact. 

Later, we use \eq{contact_eq} in MIQP and here we approximate \eq{contact_eq} as linear inequality constraints. In particular, the only constraints which need to be approximated are \eq{limit_surface_full} inside \eq{contactA}. For notation simplicity, we use $f^x, f^y, f^z$ as $x, y, z$ elements of $\mathbf{f}_t^{i, c}$, respectively and  $m^z$ as a $z$ element of  $\mathbf{m}_t^{i, c}$. 
\begin{equation}
    |f^i| \leq \mu_c f^z - \frac{|m^z|}{k}, i  = \{x, y\},  |m^z| \leq k \mu_c f^z \label{linear_ls}
\end{equation}
The issue in \eq{linear_ls} is  that we need to consider two absolute value of decision variables $|f^i|, |m^z|$ simultaneously. We employ a piece-wise linear representation with integer variables to deal with \eq{linear_ls} as follows:
\begin{subequations}
\begin{flalign}
\gamma_t^{i, c} = 1\Longrightarrow m^z_- = 0, 
\gamma_t^{i, c} = 0\Longrightarrow m^z_+ = 0\label{ls_int1}\\
    |f^i| \leq \mu_c f^z - \frac{m^z_+}{k}, |f^i| \leq \mu_c f^z - \frac{m^z_-}{k}, i  = \{x, y\},  \\
    m^z = m^z_+ -  m^z_-, m^z_+ \geq 0, m^z_- \geq 0, |m^z| \leq k \mu_c f^z \label{ds}
\end{flalign}
\label{linear_frictional_LS}
\end{subequations}
where $m^z_+, m^z_-$ are non-negative values and are the moment along  $z$-axis in $\Sigma_{c}$ in the positive and negative direction. Using \eq{ls_int1}, we decompose \eq{linear_ls} into two inequality constraints in the positive and negative direction of $m^z$. 

\subsubsection{Wrench Transformation}
The wrench in $\Sigma_W$ can be obtained from local wrenches:
\begin{equation}
        \boldsymbol{\lambda}_{t}^i = \sum_{c=1}^{C}  {^{W}_{c}T} \mathbf{f}^{i, c}_{t}, \ \boldsymbol{\tau}_{t}^i = \sum_{c=1}^{C}  {^{W}_{c}T} \mathbf{m}^{i, c}_{t}, 
        \label{trans}
\end{equation}
With constraints \eq{contact_eq}, \eq{trans} converts a specific local wrench where the finger makes contact to the wrenches in $\Sigma_W$. 

\subsubsection{Collision-Avoidance}
The constraints in \eq{contact_eq} could allow the fingers to penetrate into the holds. 
To avoid the penetration, the collision-avoidance constraints are given by:
\begin{subequations}
\begin{flalign}
\beta_t^{i, v, h} = 0 \Longrightarrow
 {^{v_h}_{W}T}\mathbf{p}_t^i \cdot \mathbf{n}^{v_h} \leq s^{v_h}, \forall h = 1, \ldots, n_v  \label{collisionA}
\\
\sum_{h=1}^{n_v} \beta_t^{i, v, h} \leq n_v - 1 \label{sum_collsion}
\end{flalign}
\label{collision_eq}
\end{subequations}
where $\mathbf{n}^{v_h}$ is the normal vector to face $h$ of obstacle $v$ in $\Sigma_{v_h}$ and $s^{v_h}$ is a scalar to decide the location of the plane. 
${^{v_h}_{W}T}\mathbf{p}_t^i$ represents $\mathbf{p}_t^i$ in $\Sigma_{v_h}$.
\eq{collision_eq} means that the finger needs to be outside at least one face of the obstacle.

%

\subsection{Alternating Direction Method of Multipliers (ADMM)}
%
ADMM solves the optimization problem with consensus constraints as follows: 
\begin{equation}
\min _{\boldsymbol{\eta}, \boldsymbol{\delta}} f(\boldsymbol{\eta}) + g(\boldsymbol{\delta}), \
\text{s. t. } A\boldsymbol{\eta} + B\boldsymbol{\delta} = \mathbf{c}
\label{admm0}
\end{equation}
where $\boldsymbol{\eta} \in \mathbb{R}^{n_\eta}, \boldsymbol{\delta} \in \mathbb{R}^{n_\delta}, A \in \mathbb{R}^{n_\epsilon \times n_\eta}, B \in \mathbb{R}^{n_\epsilon \times n_\delta}, \mathbf{c} \in \mathbb{R}^{n_\epsilon}$. 
By decomposing the original
problem with two smaller-scale problems and solving each problem with considering consensus, ADMM effectively solves the original optimization problem with faster convergence \cite{MAL-016}. 


The augmented Lagrangian of \eq{admm0} can be given by:
\begin{equation}
\begin{aligned}
L_{\rho}(\boldsymbol{\eta}, \boldsymbol{\delta}, \boldsymbol{\epsilon}) = f(\boldsymbol{\eta})+g(\boldsymbol{\delta})  + \frac{\rho}{2}(\|A \boldsymbol{\eta}+B \boldsymbol{\delta}-\mathbf{c} + \boldsymbol{\epsilon}\|_{2}^{2})
\label{scaled_admm}
\end{aligned}
\end{equation}
with $\rho > 0$. $\boldsymbol{\epsilon} \in \mathbb{R}^{n_\epsilon}$ is the dual variable associated with the constraints $A\boldsymbol{\eta} + B\boldsymbol{\delta} = \mathbf{c}$. Then, ADMM finds the solution by taking the following steps recursively:
\begin{subequations}
\begin{flalign}
&\boldsymbol{\eta}^{k+1}:= \argmin_{\boldsymbol{\eta}} L_{\rho}\left(\boldsymbol{\eta}, \boldsymbol{\delta}^{k}, \boldsymbol{\epsilon}^{k}\right) \label{admm00}\\
&\boldsymbol{\delta}^{k+1}:= \argmin_{\boldsymbol{\delta}} L_{\rho}\left(\boldsymbol{\eta}^{k+1}, \boldsymbol{\delta}, \boldsymbol{\epsilon}^{k}\right)\label{admm11} \\
&\boldsymbol{\epsilon}^{k+1}:=\boldsymbol{\epsilon}^{k}+A \boldsymbol{\eta}^{k+1}+B \boldsymbol{\delta}^{k+1}- \mathbf{c} \label{consensus_admm}
\end{flalign}
\label{2-block-admm}
\end{subequations}
%
%
The natural extension of the  two-block ADMM is
$I$-block ADMM: 
\begin{equation}
\min_{^{1}\boldsymbol{\eta}, ^{2}\boldsymbol{\eta}, \ldots, ^{I}\boldsymbol{\eta}} \sum_{i=1}^{I} f_{i}\left({^{i}\boldsymbol{\eta}}\right) \
\text{s. t. } ^{i}_j{\eta} = {^{{\mathcal{G}(i, j)}}\delta},  \forall (i, j) \in \mathcal{G}
\label{consensus}
\end{equation}
where $^{i}_j{\eta}$ is the $j$-th local decision variable of $i$-th block optimization problem. 
$^{{g}}\delta = {^{{\mathcal{G}(i, j)}}\delta}$ is the $h$-th element of the global decision variable, $\boldsymbol{\delta} \in \mathbb{R}^{G}$. 
$\mathcal{G}$ is a bipartite graph formed from global decision variables and local decision variables where each edge represents a consensus constraint (see \cite{MAL-016}). 
We also denote $\mathcal{N}(g) = \{i, j|(i, j) \in \mathcal{G}\}$ as the set of all local variables connected to $^g\delta$. 
Denote $^i_j\epsilon$ as the dual variable associated with the consensus constraints.
The multi-block consensus problem can be solved as follows:
$
\begin{aligned}
^{i}\boldsymbol{\eta}^{k+1}:= \argmin_{^{i}\boldsymbol{\eta}} f_i(^{i}\boldsymbol{\eta}) + \sum_{g \in \mathcal{G}(i, j)}\frac{\rho_i}{2}\|^i_j\eta - {^g\delta^k}+ {^i_j\epsilon^k}\|_{2}^{2}, \label{x_iopt} \\
^g\delta^{k+1}:= \argmin_{^g\delta}  \sum_{i, j \in \mathcal{N}(g)}\frac{\rho_i}{2}\|^i_j\eta^{k+1} {-} {^g\delta}+ {^i_j\epsilon^k}\|_{2}^{2}, \label{y_iopt} \\
^i_j\epsilon^{k+1}:=  {^i_j\epsilon^{k}} + (^i_j\eta^{k+1} {-} {^g\delta}^{k+1}),  \forall (i, j) \in \mathcal{G}
\label{admm1}
\end{aligned}
$

\begin{figure}
    \centering
    \includegraphics[width=0.3\textwidth, clip]{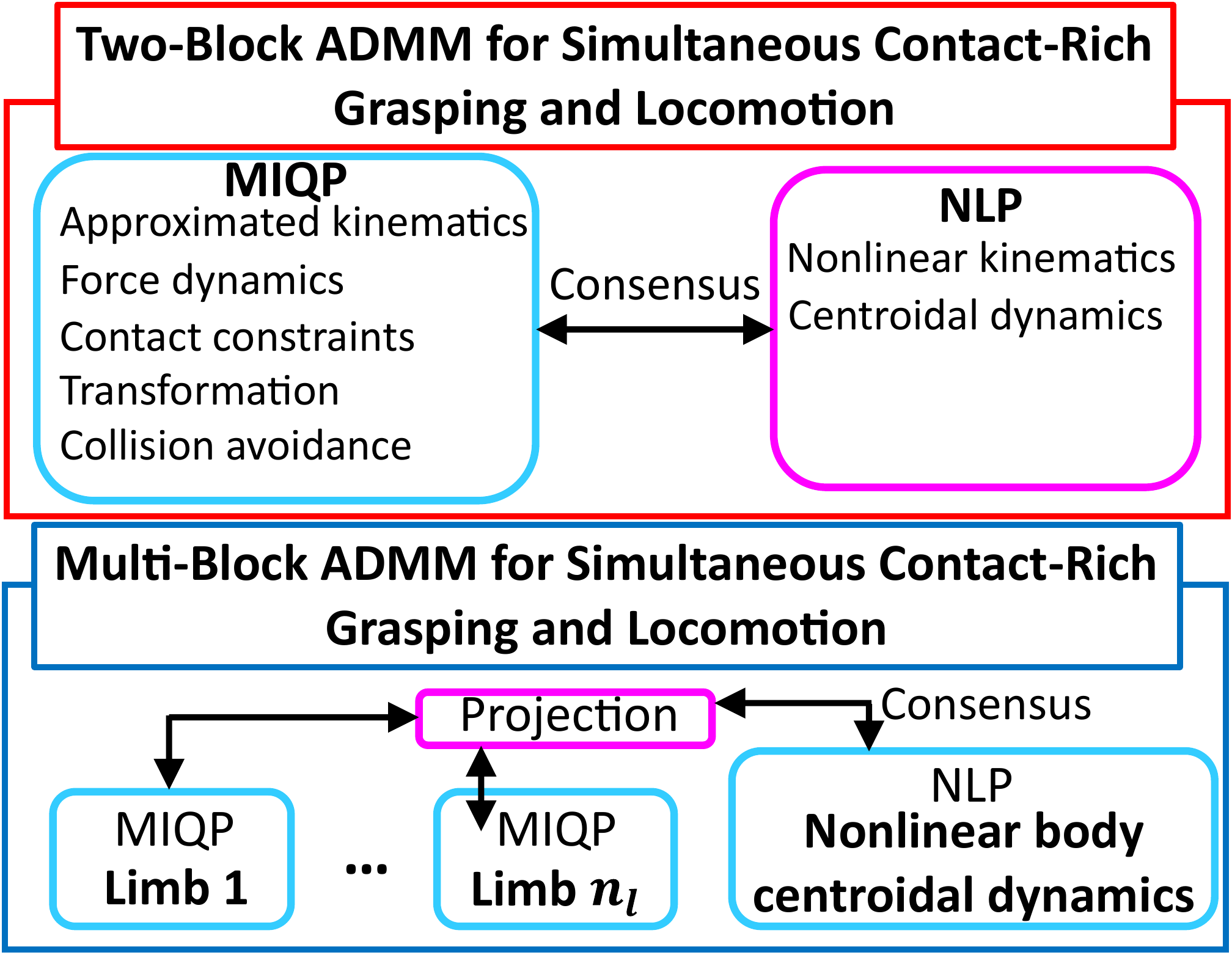} %
    \caption{We propose two distributed optimization based on ADMM specific for motion planning of limbed robots. (top): Two-block ADMM where MIQP considers discrete constraints and NLP considers nonlinear constraints so that the planner effectively solves the MINLP once these two optimization problems achieve  consensus. (bottom): Multi-block ADMM  where the planner consists of $n_l$ MIQP problems and one NLP.}
    \label{fig:consensus_figure}
\end{figure}
\subsection{Distributed Optimal Control Problem}\label{DOCP}
We propose the following distributed optimization framework so that the original intractable MINLP problem (P1) becomes tractable.
The key idea is that we use an NLP solver to solve NLP with nonlinear continuous constraints  and a MIP solver to solve MIQP with discrete constraints  so that we can employ the strength of each solver. 
%
Our proposed distributed optimization from P1 solves the following optimization problems recursively (\textbf{P2}):
\begin{subequations}
\begin{flalign}
\textbf{P2a}: \min_{\boldsymbol{\eta}, y, z} \sum_{t=0}^{N-1}J(\boldsymbol{\eta}^k_t) + \frac{\rho}{2}\|\boldsymbol{\eta}_t^k-\boldsymbol{\delta}_t^k + \boldsymbol{\epsilon}_t^k\|_{2}^{2}  \\ 
\begin{aligned}
\text{s.t. } \{ \forall t,  \eq{bounds_eq}, \eq{force_dynamics} \} \bigcap \{ \forall t, i, \eq{kinematics_eq} \} \\  \bigcap \{ \forall t, i, c, \eq{contact_eq}, \eq{trans} \}\bigcap \{ \forall t, i, v, \eq{collision_eq} \} 
\end{aligned}\\
%
%
\textbf{P2b}: \min_{\boldsymbol{\delta}}  \sum_{t=0}^{N-1} \frac{\rho}{2}\|\boldsymbol{\eta}_t^{k+1}-\boldsymbol{\delta}_t^k + \boldsymbol{\epsilon}_t^k\|_{2}^{2}  \\  \text{s.t. } \{ \forall t,  \eq{bounds_eq}, \eq{Centroidal_eq} \}  \bigcap \{ \forall t, i, \eq{kinematics_eq} \} \\
\textbf{P2c}:    \boldsymbol{\epsilon}^{k+1}_t \longleftarrow \boldsymbol{\epsilon}^{k}_t+\boldsymbol{\eta}^{k+1}_t- \boldsymbol{\delta}^{k+1}_t
\end{flalign}
\label{admm_P2b}
\end{subequations}
%
%
%
where $\boldsymbol{\eta}^k_t = [x_k^\top, u_k^\top]^\top, \boldsymbol{\eta}^k = [\boldsymbol{\eta}^{k\top}_{0}, \ldots, \boldsymbol{\eta}^{k\top}_N]^\top$.
The indexes in \eq{admm_P2b} are $t = 0, \ldots, N-1$, $i = 1, \ldots, n_f$, $c = 0, \ldots, C$, $v = 1, \ldots, V$.
We create copies of variables $\boldsymbol{\eta}_t$ as $\boldsymbol{\delta}_t$ and $\boldsymbol{\delta}^k = [\boldsymbol{\delta}^{k\top}_{0}, \ldots, \boldsymbol{\delta}^{k\top}_N]^\top$. 
P2a is MIQP. P2b is continuous NLP. Since both MIQP and NLP can be efficiently solved using off-the-shelf solvers, our proposed method would converge earlier than the method solving P1 using MINLP solvers. 
Also, our method considers all constraints from grasping, locomotion, and contact once it achieves the consensus between MIQP and NLP. Thus, it does not suffer from the infeasible issue of hierarchical planning explained in Sec~\ref{sec:related_work}.
%

\textit{Remark 1}: In P2, we do not consider the consensus of $\mathbf{y}, \mathbf{z}$. For $\mathbf{y}$, $\mathbf{u}$ indirectly has an effect on $\mathbf{y}$ via \eq{trans} so we do not explicitly enforce consensus constraints for $\mathbf{y}$, which enables more efficient ADMM computation. For $\mathbf{z}$, because it is quite difficult for NLP to satisfy discrete constraints, we do not enforce consensus between P2a and P2b.

\subsection{Multi-Block Distributed Optimal Control Problem}\label{multi-admm-climbing}
We propose another option to solve the MINLP which does not involve many difficult constraints (e.g., discrete constraints) based on multi-block ADMM as follows (\textbf{P3}):
$
\begin{aligned}
\textbf{P3a}: \min_{^i\boldsymbol{\eta}, y, z} \sum_{t=0}^{N-1} \left( {^i}J(^i\boldsymbol{\eta}^k_t)  + \sum_{g \in \mathcal{G}(i, j)}\frac{\rho_i}{2}\|^i_j\eta_t - {^g\delta^k_t}+ {^i_j\epsilon^k_t}\|_{2}^{2}\right),  \\ 
\text{s.t. } \text{for MIQP: P2a}, \text{for NLP: P2b}, \\
\textbf{P3b (projection)}: \min_{^g\delta}  \sum_{t=0}^{N-1} \sum_{i, j \in \mathcal{N}(g)}\frac{\rho_i}{2}\|^i_j\eta^{k+1}_t {-} {^g\delta}_t+ {^i_j\epsilon^k_t}\|_{2}^{2}\\
\textbf{P3c}: {^i_j}\epsilon^{k+1}:=  {{^i_j}\epsilon^{k}_t} + (^i_j\eta^{k+1}_t {-} {^g\delta}^{k+1}_t)
\end{aligned}
$ 
where $^i\boldsymbol{\eta}^k_t = [^ix_k^\top, {^i}u_k^\top]^\top$.
We run P3a with constraints from P2a for each limb to solve discrete constraints, resulting in total $n_l$ MIQP problems in parallel. We run one P3a with constraints from P2b to solve nonlinear constraints. Then, we run P3b as projection (see \fig{fig:consensus_figure}).

\section{Experimental Results}\label{sec:results}
In this section, we validate our proposed methods for two tasks: walking and free-climbing. Through the experiments, we try to answer the following questions:
\begin{enumerate}
\item Can our proposed optimization generate open-loop trajectories efficiently? 
\item Can our proposed formulation consider patch contacts with micro-spines explicitly?
\item How do the generated trajectories behave in a real four-limbed robot?
\end{enumerate}

\subsection{Implementation Details}
For optimization settings, we implemented our method using Gurobi  \cite{gurobi} for solving MIQP  and ipopt \cite{wachter2006implementation} with PYROBOCOP \cite{Raghunathan2022} for solving NLP. The optimizations are done on a computer with the Intel i7-8565U. The trajectories discussed in this section are from two-block ADMM and we use the solution from MIQP (P2a).

We implemented the results of our proposed methods on a real four-limbed robot \cite{iros22_scaler} equipped with two-finger grippers \cite{9635872}.    The grippers make patch contact with micro-spines, which are mechanically constrained such that the patch is always perpendicular to the surface during contact. This satisfies one of our assumptions in Sec~\ref{section_microspine}. Each limb consists of 7 DoF, where 6 DoF are to actuate the joints of the limb and 1 DoF is to actuate the gripper. The robot weighs 9.6 kg.  The admittance control was used to track the reference wrenches from the planner \cite{iros22_admittance}. 
Free-climbing experiments were conducted on a rugged wall with gradient  $\ang{45}$.
Hardware experiments can be viewed in the accompanying video\footnote{https://youtu.be/QLH1shghqQ0}. 

\textit{Remark 2}: We set $dt$ to the  large value and have tight bounds on $\dot{\mathbf{p}}_t^i$ for   free-climbing, resulting in slow motions in hardware. This is because our robot uses linear actuators to actuate fingers, whose internal motor velocity is slow.

\subsection{Computation Results}\label{comp_result}
\subsubsection{Convergence Analysis}\label{conv_analysis}
We discuss the convergence of our ADMMs. For walking with point contacts on a flat plane without, we set $dt=0.08$, $N=40$, $\rho=1.5$, $\mu=0.6$. 
For walking, we consider point contacts (i.e., no fingers). 
Since we consider a single flat plane walking without obstacles,  we set $C= 1, V = 0$. 
For climbing, we set $dt=2.0$, $N=40$, $\rho=10$, $\mu=2.2$, $r_p=0.02$ m. This scenario considers four climbing holds which consists of five faces so we set $C = 20$, $V=4$. Both cases run ADMM for 10 iterations. 

\begin{figure}
    \centering
    \includegraphics[width=0.4\textwidth]{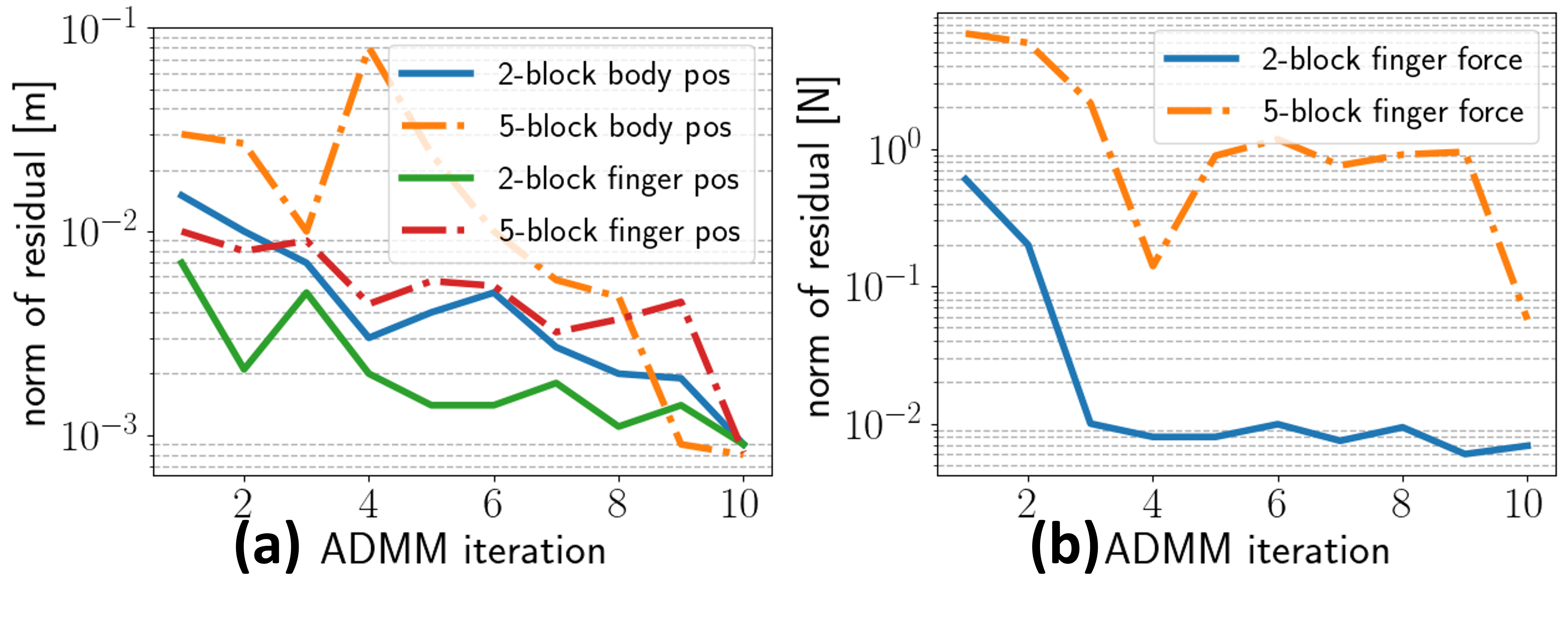} %
    \caption{Evolution of residuals for walking  using two- and five-block ADMM.  (a): Residual of body and finger positions, (b): residual of  finger forces.}
    \label{fig:walking_converge}
\end{figure}

\begin{figure}
    \centering
    \includegraphics[width=0.4\textwidth]{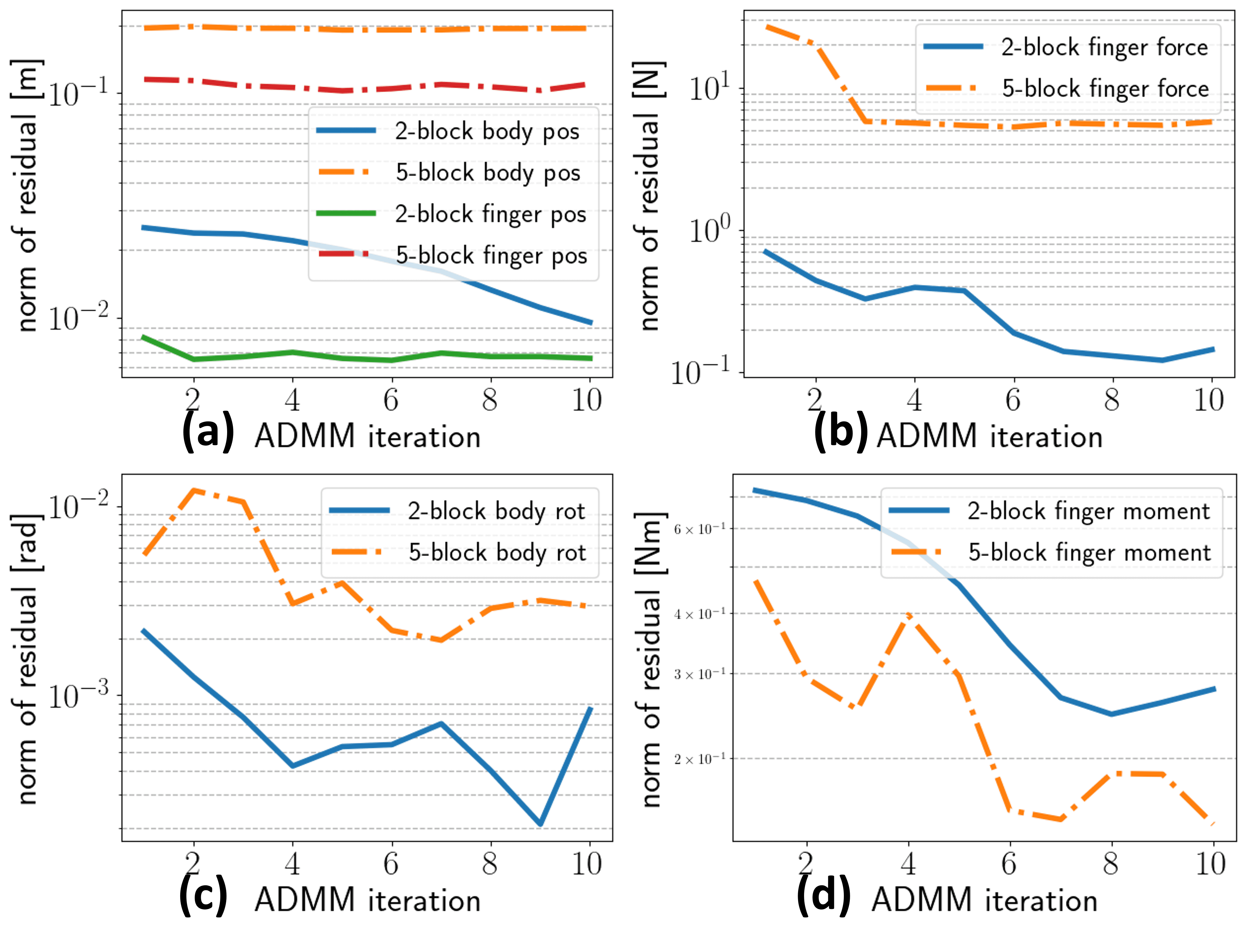} %
    \caption{Evolution of residuals for free-climbing  using two- and five-block ADMM. We show residual of (a): body  and finger positions, (b): finger forces, (c): body rotation,  (d): finger moments.}
    \label{fig:climbing_converge}
\end{figure}

We show the evolution of residual for walking and free-climbing using our two-  and five-block ADMM in \fig{fig:walking_converge} and \fig{fig:climbing_converge}. 
Overall, we observe that our proposed ADMM converges with small enough norms of residual.
We also observe that for free-climbing, our two-block ADMM shows faster convergence than our five-block ADMM and for walking, both two- and five-block ADMM shows similar convergence. This is because the problem is so complicated that it is quite difficult for five-block ADMM to achieve consensus,  
resulting in higher norms of residual. In contrast, our two-block ADMM only needs to achieve consensus for two optimization problems, resulting in  lower norms of residuals.

\begin{figure}[t!]
    \centering
    \includegraphics[width=0.499\textwidth]{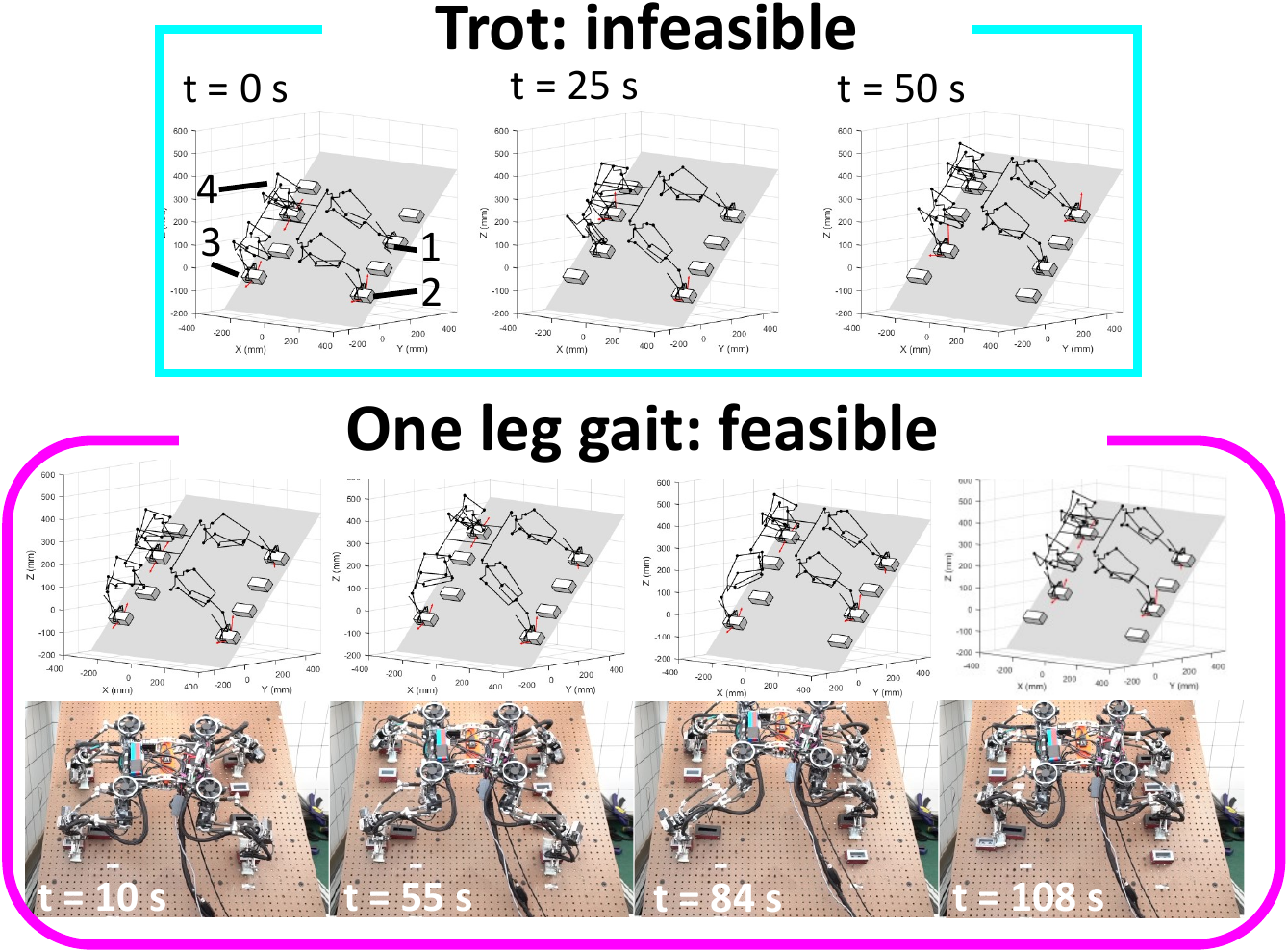} 
    \caption{Change of modes as our ADMM proceeds with snapshots of hardware experiments. (top): after 1 iteration of our ADMM, the planner generates a trot gait, which is physically infeasible since MIQP is not fully influenced by nonlinear constraints yet. (bottom): after 6 iterations, the planner finds a physically feasible one leg gait sequence ($1\rightarrow4\rightarrow2\rightarrow3$).}
    \label{fig:top_figure}
\end{figure}

We also discuss the generated trajectories for free-climbing by our two-block ADMM as shown in \fig{fig:top_figure}. The trot gait trajectory is the result of our ADMM after 1 iteration. This trot gait is physically infeasible (i.e., tumbling) since MIQP is not aware of nonlinear moment constraints \eq{moment_dynamics}. After 6 iterations of our ADMM, our planner generates a physically feasible one leg gait. Because ADMM converges, MIQP is now aware of \eq{moment_dynamics} so that the generated trajectory does not make the robot tumble anymore, resulting in physically feasible trajectories.
%
We do not observe this mode change for the walking task since the walking task is naturally more stable than the free-climbing task so our ADMM does not face the need for mode change. 

\subsubsection{Computation Time, Success Rate}\label{compt_time}
We compare our ADMM with the benchmark optimization using NLP with respect to the computation time and the success rate of finding a feasible solution for walking and climbing problems. As a benchmark, we solve P1 as NLP using a technique in \cite{STEIN20041951}. We sample ten feasible initial and terminal states and calculate the average computation time and the success rate where the optimization could find a solution. We use the same parameters in Sec~\ref{conv_analysis}.
\begin{table}[t]
    \caption{{Comparison of computation time and success rate for walking and climbing problems between benchmark optimization based on P1 using \cite{STEIN20041951}, our two-block ADMM, and our multi-block ADMM. For ADMMs, the computation time is calculated as  the total computation time until the norm of residual for body  and foot positions  converge to  $0.03$ m and the norm of residual for reaction forces converge to within $0.5$ N. For computation time, we show the mean time and 99 $\%$ confidence interval.}} 
    \centering
    \begin{tabular}{c|c|c}
    Walking & Computation time [s]& Success rate [$\%$]  \\
         \hline\hline Benchmark & 3275  & 10\\
         \hline Our two-block ADMM & 168 $\pm$ 50 & 100 \\
          \hline Our five-block ADMM  & 44 $\pm$ 21 & 100 \\
             Climbing & &  \\
         \hline\hline Benchmark  & N/A & 0\\
         \hline Our two-block ADMM & 619  $\pm$ 99 & 100 \\
         \hline Our five-block ADMM  & 970 $\pm$ 60 & 90
    \end{tabular}
    \label{walking_table}
\end{table} 
\tab{walking_table} shows
that our ADMMs show smaller computation time and a  higher success rate against the benchmark method.
For walking, our five-block ADMM shows smaller computation time and for free-climbing, our two-block ADMM shows smaller computation time.
In other words, if the problem is not so complicated (e.g., walking), our five-block ADMM can be an option to solve the problem even though it takes more iteration to converge. This is because our five-block ADMM spends less time for each iteration so that the total computation time can be  small. 

Our proposed ADMM does not employ a warm-start. This is because each block on our ADMM solves a smaller-scale optimization problem, which is less sensitive to initial guesses compared to larger-scale complicated optimization problems (e.g., MINLP).
We did not observe a significant reduction of computation time with our warm-start, but designing warm-start for consensus constraints would be 
promising. 




\subsection{Results of Our Generated Trajectories}\label{result_tra_to}
\begin{figure}
    \centering
    \includegraphics[width=0.499\textwidth]{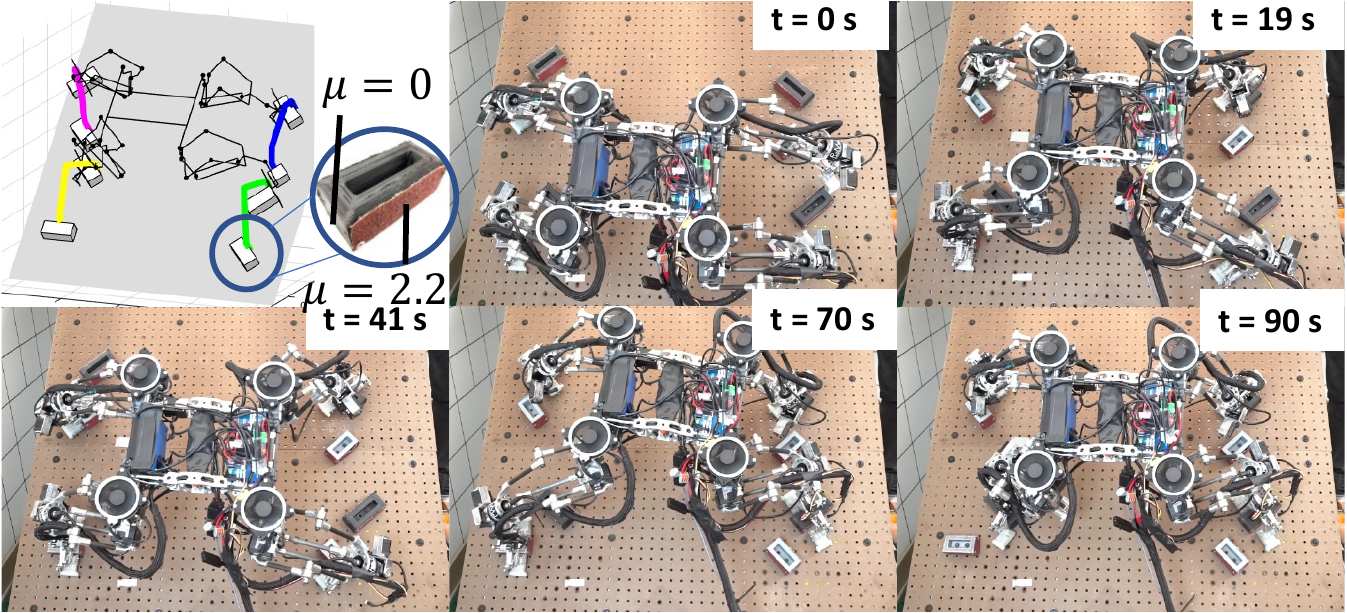} %
    \caption{Our planned trajectories on holds with varying coefficients of frictions. The robot grasps the faces whose coefficients of friction are high.}
    \label{fig:rotation_climb}
\end{figure}

\subsubsection{Collision-Avoidance}
This scenario focuses on environments with obstacles.  
We set $dt=2.0$, $N=120$, $\rho=15$, $\mu=2.2$, $r_p=0.02$ m. We consider 16 climbing holds and 3 obstacles and
 run our two-block ADMM for 3 iterations. 

The generated trajectory with snapshots of the hardware experiment is shown in \fig{fig:collision}. Our ADMM successfully generates collision-free trajectories under a number of discrete constraints. Here we discuss three points in \fig{fig:collision}. Around point A, since the height of the obstacle is not so high, our ADMM generates the trajectory that gets over the obstacle. Around point B, since the height of the obstacle is high, the robot cannot get over the obstacle. Thus, our ADMM generates the trajectory that takes a detour around the obstacle.
Around point D, limb 4 could directly make contact on D from C, but due to the obstacle, the robot first makes a contact on E and then goes to D. 
\subsubsection{Slippery Rotated Holds}
This scenario shows that our ADMM designs trajectories with varying coefficients of friction on rotated holds. We set $dt=2.0$, $N=40$, $\rho=15$, $r_p=0.02$ m. The environment consists of 8 climbing holds with different orientations along $z$-axis in $\Sigma_W$. In \fig{fig:rotation_climb}, each climbing hold has a $\mu$ where the front and back faces are covered by 36-sand papers ($\mu = 2.2$), and the left and right faces are covered by the material with $\mu = 0$. 

The generated trajectory with snapshots of  experiments is shown in \fig{fig:rotation_climb},  where the robot grasps the front or back face (high friction). To grasp the front or back face,  the robot rotates the grippers so that the fingers make contacts perpendicular to the faces. In short, our planner could find feasible trajectories subject to the pose, wrenches, and contacts together. In contrast, hierarchical planners may only find the infeasible solution  since it cannot consider coupling constraints in general. 


\subsection{Contact Modeling Results}
\subsubsection{Results of Micro-Spine Limit Surface}
\begin{figure}
    \centering
    \includegraphics[width=0.2\textwidth]{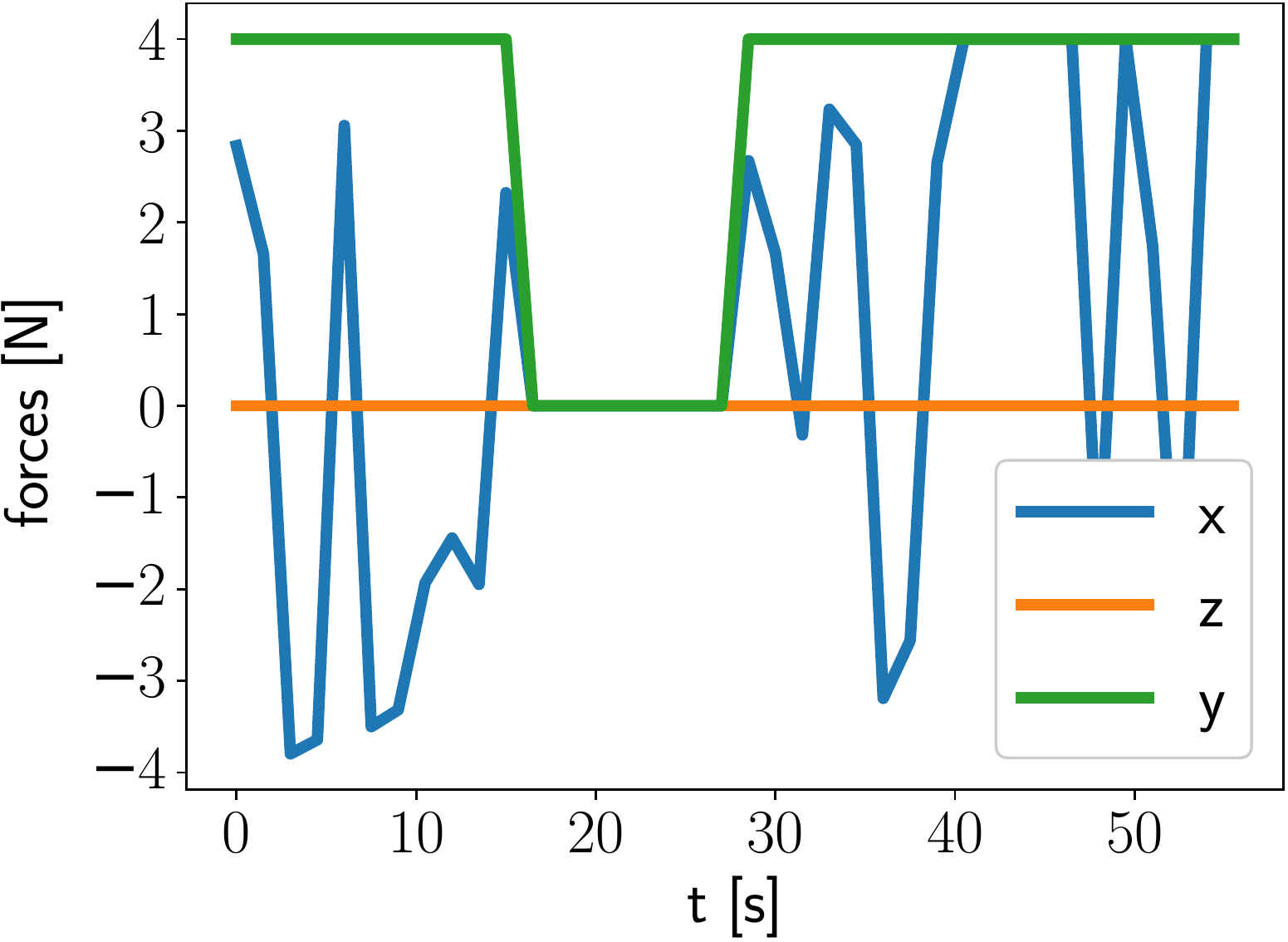} %
    \caption{Time history of reaction forces for one finger. Here, we enforce $z$ element of reaction force in $\Sigma_{C_c}$  is always zero with $f^i_\text{max} = 4$ N in \eq{limit_surface_spine}.}
    \label{fig:spine_TO}
\end{figure}

We force one of the fingers to have zero normal forces during contact (i.e., $z$ element of  $\mathbf{f}^{i, c}_t$ is set to zero for all $c = 1, \ldots, C$). Since \eq{limit_surface_spine} enables the planner to generate non-zero shear forces even under zero normal forces, we expected that our planner can still find feasible solutions for free-climbing. The result is illustrated in \fig{fig:spine_TO}. Our planner is able to generate non-zero shear force trajectories under zero normal forces.

\subsubsection{Results of Frictional Limit Surface}
We investigate if our proposed planner can generate physically feasible trajectories under patch constraints \eq{minkosum} in hardware. 
During free-climbing, the loading shear forces and moments exist at the tip of the finger, which can lead to instability of the contact state. We hope that considering \eq{linear_frictional_LS} counteracts these loading shear forces and moments so that the contact state is stable. 
Thus, our planner creates two different trajectories, one considering patch constraints (Traj A) and the other one not considering them (Traj B). To simplify the analysis, for both cases, the planner set the same constant shear force  and the moment ($f_x, m_z$ in \eq{linear_ls}) during contacts.
We tested on the box covered by 36-grit sandpapers  with specified  $f_x = 9.0$ N and $m_z = \{0, 0.3, 0.6\}$ Nm.  

\begin{table}[t]
    \caption{{Comparison of the number of  slipping between a  generated trajectory with patch constraints (Traj A) and without patch constraints (Traj B) over 5 samples for each case.}} 
    \centering
    \begin{tabular}{c|c|c|c}
     & $m_z = 0.0$ Nm& $m_z = 0.3$ Nm & $m_z = 0.6$ Nm \\
         \hline\hline Traj A & 1 / 5  & 0 / 5  & 0 / 5 \\
         \hline Traj B & 2 / 5  & 5 / 5  & 5 / 5  
    \end{tabular}
    \label{wrench_table}
\end{table}

\begin{figure}
    \centering
    \includegraphics[width=0.49\textwidth]{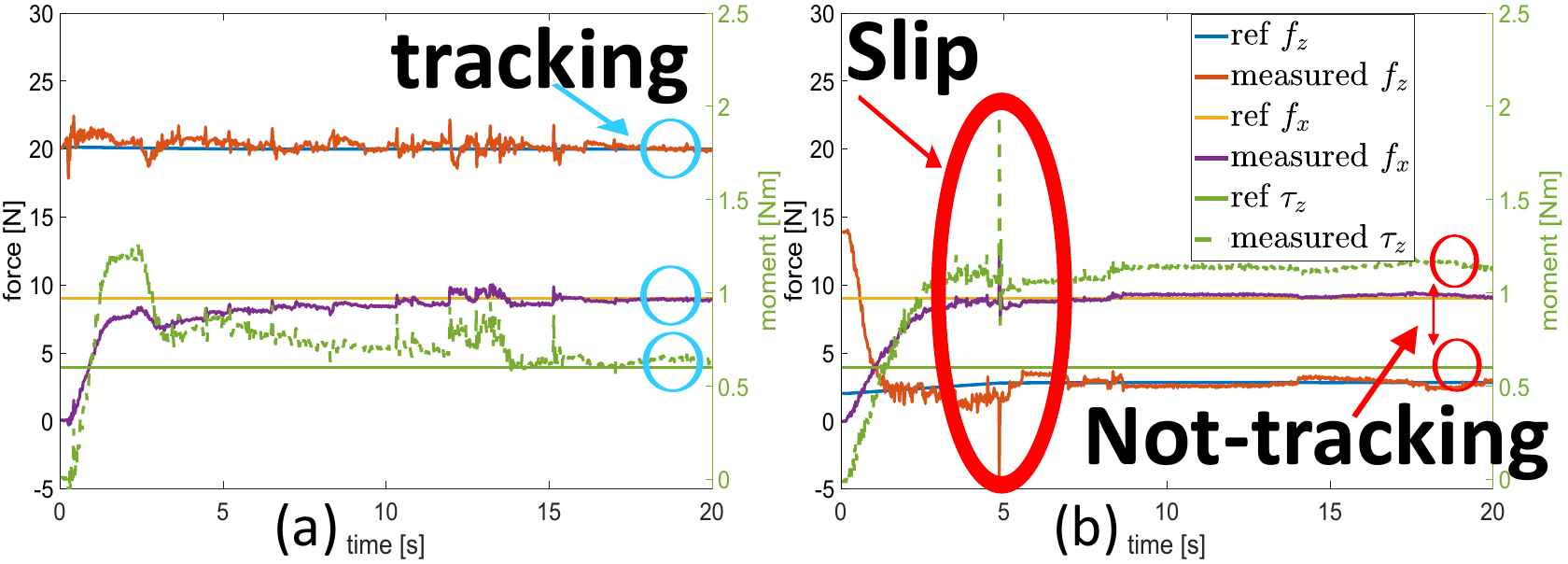} %
    \caption{Time history of wrench trajectories between the trajectory (a):  considering patch constraints and (b): not considering patch constraints. }
    \label{fig:wrench_result}
\end{figure}

\tab{wrench_table} shows the empirically obtained number of slipping for Traj A is much smaller than that for  Traj B. This is because Traj A generates higher normal forces to avoid slipping because of patch constraints. We also show the time history of Traj A and B in \fig{fig:wrench_result} given $m_z = 0.6$ Nm, $f_x = 9.0$ N settings. In \fig{fig:wrench_result} (a), since our ADMM generates larger normal forces  because of  \eq{linear_ls}, the finger does not slip and the admittance controller could track all wrenches. In contrast, in \fig{fig:wrench_result} (b), since our ADMM generates smaller normal forces, the finger slips, and the controller could not track the wrenches. Therefore, we successfully verify in hardware that considering \eq{linear_frictional_LS} helps avoid slipping for patch contacts, resulting in a more stable contact state.

\section{Conclusion and Future Work}\label{sec:discussion}
This paper presents a model-based motion planning algorithm based on distributed optimization for solving nonlinear contact-rich systems efficiently. We first propose the complete optimization formulation and extend it to our proposed distributed formulations. We also discuss the limit surface of two-finger grippers with patch contacts and micro-spines. We verify the efficiency of our proposed formulation and demonstrate the generated trajectories in hardware experiments. 

One limitation of our ADMM is that the computation is  demanding once the number of discrete constraints increases. In order to use our framework in MPC, we need to run it with a much faster runtime. 
One promising direction is to design heuristics online \cite{9134792}. We hope that  we can accelerate our framework as ADMM iteration proceeds based on previous solutions. 
Another limitation  during hardware experiments is that it is important to use accurate physical parameters. Otherwise, the robot may not be able to execute the planned trajectory. Thus, we are interested in robustifying our method so that the robot can realize the planned trajectory  under uncertain parameters~\cite{shirai2022chance}. 




\bibliographystyle{IEEEtran}
\bibliography{main.bib}

\begin{thebibliography}{10}
\providecommand{\url}[1]{#1}
\csname url@samestyle\endcsname
\providecommand{\newblock}{\relax}
\providecommand{\bibinfo}[2]{#2}
\providecommand{\BIBentrySTDinterwordspacing}{\spaceskip=0pt\relax}
\providecommand{\BIBentryALTinterwordstretchfactor}{4}
\providecommand{\BIBentryALTinterwordspacing}{\spaceskip=\fontdimen2\font plus
\BIBentryALTinterwordstretchfactor\fontdimen3\font minus
  \fontdimen4\font\relax}
\providecommand{\BIBforeignlanguage}[2]{{%
\expandafter\ifx\csname l@#1\endcsname\relax
\typeout{** WARNING: IEEEtran.bst: No hyphenation pattern has been}%
\typeout{** loaded for the language `#1'. Using the pattern for}%
\typeout{** the default language instead.}%
\else
\language=\csname l@#1\endcsname
\fi
#2}}
\providecommand{\BIBdecl}{\relax}
\BIBdecl

\bibitem{Circus}
F.~Shi \emph{et~al.}, ``Circus anymal: A quadruped learning dexterous
  manipulation with its limbs,'' in \emph{Proc. 2021 IEEE Int. Conf. Robot.
  Automat.}, 2021.

\bibitem{doi:10.1080/01691864.2012.686345}
K.~Bouyarmane and A.~Kheddar, ``Humanoid robot locomotion and manipulation step
  planning,'' \emph{Adv. Robot.}, vol.~26, no.~10, pp. 1099--1126, 2012.

\bibitem{9359455}
M.~Murooka \emph{et~al.}, ``Humanoid loco-manipulation planning based on graph
  search and reachability maps,'' \emph{IEEE Robot. Autom. Lett.}, vol.~6,
  no.~2, pp. 1840--1847, 2021.

\bibitem{8957267}
M.~P. Polverini \emph{et~al.}, ``Multi-contact heavy object pushing with a
  centaur-type humanoid robot: Planning and control for a real demonstrator,''
  \emph{IEEE Robot. Autom. Lett.}, vol.~5, no.~2, pp. 859--866, 2020.

\bibitem{9478184}
I.~Kumagai \emph{et~al.}, ``Multi-contact locomotion planning with bilateral
  contact forces considering kinematics and statics during contact
  transition,'' \emph{IEEE Robot. Autom. Lett.}, vol.~6, no.~4, pp. 6654--6661,
  2021.

\bibitem{7989643}
A.~Parness \emph{et~al.}, ``Lemur 3: A limbed climbing robot for extreme
  terrain mobility in space,'' in \emph{Proc. 2017 IEEE Int. Conf. Robot.
  Automat.}, 2017, pp. 5467--5473.

\bibitem{doi:10.1177/0278364913506757}
M.~Posa, C.~Cantu, and R.~Tedrake, ``A direct method for trajectory
  optimization of rigid bodies through contact,'' \emph{Int. J. Rob. Res.},
  vol.~33, no.~1, pp. 69--81, 2014.

\bibitem{yuki2021pivot}
Y.~Shirai \emph{et~al.}, ``Robust pivoting: Exploiting frictional stability
  using bilevel optimization,'' in \emph{Proc. 2022 IEEE Int. Conf. Robot.
  Automat.}, 2022.

\bibitem{8740889}
J.~Carius \emph{et~al.}, ``Trajectory optimization for legged robots with
  slipping motions,'' \emph{IEEE Robot. Autom. Lett.}, vol.~4, no.~3, pp.
  3013--3020, 2019.

\bibitem{8141917}
B.~Aceituno-Cabezas \emph{et~al.}, ``Simultaneous contact, gait, and motion
  planning for robust multilegged locomotion via mixed-integer convex
  optimization,'' \emph{IEEE Robot. Autom. Lett.}, vol.~3, no.~3, pp.
  2531--2538, 2018.

\bibitem{8283570}
A.~W. Winkler \emph{et~al.}, ``Gait and trajectory optimization for legged
  systems through phase-based end-effector parameterization,'' \emph{IEEE
  Robot. Autom. Lett.}, vol.~3, no.~3, pp. 1560--1567, 2018.

\bibitem{9166536}
T.~Stouraitis \emph{et~al.}, ``Online hybrid motion planning for dyadic
  collaborative manipulation via bilevel optimization,'' \emph{IEEE Trans.
  Robot.}, vol.~36, no.~5, pp. 1452--1471, 2020.

\bibitem{9113247}
Y.~Shirai \emph{et~al.}, ``Risk-aware motion planning for a limbed robot with
  stochastic gripping forces using nonlinear programming,'' \emph{IEEE Robot.
  Autom. Lett.}, vol.~5, no.~4, pp. 4994--5001, 2020.

\bibitem{nguyen2021contact}
C.~Nguyen and Q.~Nguyen, ``Contact-timing and trajectory optimization for 3d
  jumping on quadruped robots,'' \emph{arXiv preprint arXiv:2110.06764}, 2021.

\bibitem{MAL-016}
S.~Boyd \emph{et~al.}, ``Distributed optimization and statistical learning via
  the alternating direction method of multipliers,'' \emph{Foundations and
  Trends® in Machine Learning}, vol.~3, no.~1, pp. 1--122, 2011.

\bibitem{8793878}
R.~Budhiraja, J.~Carpentier, and N.~Mansard, ``Dynamics consensus between
  centroidal and whole-body models for locomotion of legged robots,'' in
  \emph{2019 Int. Conf. Robot. Automat.}, 2019, pp. 6727--6733.

\bibitem{9147887}
Z.~Zhou and Y.~Zhao, ``Accelerated admm based trajectory optimization for
  legged locomotion with coupled rigid body dynamics,'' in \emph{2020 American
  Control Conference (ACC)}, 2020, pp. 5082--5089.

\bibitem{aydinoglu2021realtime}
A.~Aydinoglu and M.~Posa, ``Real-time multi-contact model predictive control
  via admm,'' in \emph{Proc. 2022 IEEE Int. Conf. Robot. Automat.}, 2022.

\bibitem{9620665}
O.~Shorinwa and M.~Schwager, ``Distributed contact-implicit trajectory
  optimization for collaborative manipulation,'' in \emph{Proc. 2021 Int. Symp.
  Multi. Robo. Multi. Agent. Syst.}, 2021, pp. 56--65.

\bibitem{doi:10.1177/027836499601500603}
R.~D. Howe and M.~R. Cutkosky, ``Practical force-motion models for sliding
  manipulation,'' \emph{Int. J. Rob. Res.}, vol.~15, no.~6, pp. 557--572, 1996.

\bibitem{wang2017design}
S.~Wang, H.~Jiang, and M.~R. Cutkosky, ``Design and modeling of
  linearly-constrained compliant spines for human-scale locomotion on rocky
  surfaces,'' \emph{Int. J. Rob. Res.}, vol.~36, no.~9, pp. 985--999, 2017.

\bibitem{8416785}
K.~Hauser, S.~Wang, and M.~R. Cutkosky, ``Efficient equilibrium testing under
  adhesion and anisotropy using empirical contact force models,'' \emph{IEEE
  Trans. Robot.}, vol.~34, no.~5, pp. 1157--1169, 2018.

\bibitem{gurobi}
\BIBentryALTinterwordspacing
{Gurobi Optimization, LLC}, ``{Gurobi Optimizer Reference Manual},'' 2021.
  [Online]. Available: \url{https://www.gurobi.com}
\BIBentrySTDinterwordspacing

\bibitem{wachter2006implementation}
A.~W{\"a}chter and L.~T. Biegler, ``On the implementation of an interior-point
  filter line-search algorithm for large-scale nonlinear programming,''
  \emph{Mathematical programming}, vol. 106, no.~1, pp. 25--57, 2006.

\bibitem{Raghunathan2022}
A.~Raghunathan \emph{et~al.}, ``Pyrobocop: Python-based robotic control and
  optimization package for manipulation,'' in \emph{Proc. 2022 IEEE Int. Conf.
  Robot. Automat.}, 2022.

\bibitem{iros22_scaler}
Y.~Tanaka \emph{et~al.}, ``Scaler: A tough versatile quadruped free-climber
  robot,'' in \emph{Proc. 2022 IEEE/RSJ Int. Conf. Intell. Rob. Syst.}, 2022.

\bibitem{9635872}
------, ``An under-actuated whippletree mechanism gripper based on
  multi-objective design optimization with auto-tuned weights,'' in \emph{Proc.
  2021 Int. Conf. Intell. Rob. Syst.}, 2021, pp. 6139--6146.

\bibitem{iros22_admittance}
A.~Schperberg \emph{et~al.}, ``Auto-calibrating admittance controller for
  robust motion of robotic systems,'' \emph{arXiv preprint arXiv:2207.01033},
  2022.

\bibitem{STEIN20041951}
O.~Stein, J.~Oldenburg, and W.~Marquardt, ``Continuous reformulations of
  discrete–continuous optimization problems,'' \emph{Comp. Chem. Eng.},
  vol.~28, no.~10, pp. 1951--1966, 2004.

\bibitem{9134792}
T.~Marcucci and R.~Tedrake, ``Warm start of mixed-integer programs for model
  predictive control of hybrid systems,'' \emph{IEEE Trans. Auto. Cont.},
  vol.~66, no.~6, pp. 2433--2448, 2021.

\bibitem{shirai2022chance}
Y.~Shirai \emph{et~al.}, ``Chance-constrained optimization in contact-rich
  systems for robust manipulation,'' \emph{arXiv preprint arXiv:2203.02616},
  2022.

\end{thebibliography}

\end{document}